\documentclass[conference]{IEEEtran}
\usepackage[utf8]{inputenc}
\usepackage{cite}
\usepackage{url}            
\usepackage{amsmath,amssymb,amsfonts}
\usepackage{algorithmic}
\usepackage{graphicx}
\graphicspath{{figures/}}
\usepackage{siunitx}
\usepackage{stfloats}
\usepackage{textcomp}
\usepackage{booktabs}
\usepackage{xcolor}
\usepackage[multiple]{footmisc}
\definecolor{caption}{RGB}{0, 255, 150}
\definecolor{drawing}{RGB}{255, 0, 100}
\definecolor{floating-word}{RGB}{0, 255, 25}
\definecolor{glyph}{RGB}{125, 0, 200}
\definecolor{image}{RGB}{255, 0, 0}
\definecolor{paragraph}{RGB}{0, 255, 255}
\definecolor{page-background}{RGB}{50, 50, 50}
\definecolor{table}{RGB}{0, 100, 25}
\definecolor{title}{RGB}{0, 100, 200}
\definecolor{gt}{RGB}{0, 255, 0}
\definecolor{illustration}{RGB}{229, 102, 102}
\definecolor{text}{RGB}{102, 229, 229}
\def\BibTeX{{\rm B\kern-.05em{\sc i\kern-.025em b}\kern-.08em
    T\kern-.1667em\lower.7ex\hbox{E}\kern-.125emX}}

\begin{document}

\title{docExtractor: An off-the-shelf\\historical document element extraction}

\author{\IEEEauthorblockN{Tom Monnier and Mathieu Aubry}
  \IEEEauthorblockA{\textit{LIGM, École des Ponts, Univ Gustave Eiffel, CNRS, 
    Marne-la-vallée, France}\\
\texttt{\{tom.monnier,mathieu.aubry\}@enpc.fr}}}

\maketitle

\begin{abstract}
  We present \textit{docExtractor}, a generic approach for extracting visual elements such as 
  text lines or illustrations from historical documents without requiring any real data 
  annotation. We demonstrate it provides high-quality performances as an off-the-shelf system 
  across a wide variety of datasets and leads to results on par with state-of-the-art when 
  fine-tuned. We argue that the performance obtained without fine-tuning on a specific 
  dataset is critical for applications, in particular in digital humanities, and that the 
  line-level page segmentation we address is the most relevant for a general purpose element 
  extraction engine. We rely on a fast generator of rich synthetic documents and design a 
  fully convolutional network, which we show to generalize better than a detection-based 
  approach. Furthermore, we introduce a new public dataset dubbed \textit{IlluHisDoc} 
  dedicated to the fine evaluation of illustration segmentation in historical documents.
\end{abstract}


\section{Introduction}

In the context of a rising interest in digital humanities, the need for easy-to-use and 
efficient tools to automatically analyse document images has dramatically increased.  Yet, 
document analysis is usually broken down into multiple sub-problems depending on the precise 
task (paragraph-level page segmentation, text line detection, photograph or illumination 
extraction, etc.) and the specific type of document (modern or historical, printed or 
handwritten, its language, its condition, etc.), each often treated independently and 
requiring a specific set of training data and annotations. It is thus difficult for 
non-specialists to find their way to the suited solution and universal engines that can 
tackle multiple tasks across various types of documents would be highly beneficial.

With the rise of deep learning, impressive improvements have been made in the document 
analysis domain. Neural-based methods not only have set new state-of-the-art in most of 
document layout analysis tasks, but also enabled the development of powerful generic 
solutions that can tackle multiple analysis tasks with a same core method. Nonetheless, each 
task specific solution can hardly be used off-the-shelf as it always requires a dedicated 
training phase, involving a considerable amount of annotations and some expertise.

In this work, we tackle the problem of document element extraction as a unified line-level 
page segmentation task. We present a fast and scalable synthetic document generation engine 
that produces a wide diversity of documents with fine-grained ground truth. We show that a 
fully convolutional network trained on resulting dataset called \textit{SynDoc} (i) is a 
powerful off-the-shelf system with remarkable performances across multiple layout analysis 
tasks and (ii) leads to state-of-the-art results when fine-tuned with real data. In a 
detailed ablation study, we demonstrate that our new data generation process as well as our 
proposed network architecture are key components for these results. To better evaluate 
generalization, we also introduce a new public test dataset dubbed \textit{IlluHisDoc} and 
dedicated to the evaluation of illustration segmentation methods for historical documents.

Synthetic generation pipeline, network implementation and IlluHisDoc dataset are all 
available at our project webpage: \texttt{\small http://imagine.enpc.fr/\texttildelow 
monniert/docExtractor/}.

\section{Related work}

\textbf{\textit{Page segmentation.}}
Also called document layout analysis, page segmentation is an active research area with 
numerous competitions~\cite{gaoICDAR2017CompetitionPage2017, 
simistiraICDAR2017CompetitionLayout2017, clausnerICDAR2019CompetitionRecognition2019a, 
clausnerICDAR2019CompetitionRecognition2019} and 
datasets~\cite{antonacopoulosRealisticDatasetPerformance2009, clausnerENPImageGround2015, 
yangLearningExtractSemantic2017, zhongPubLayNetLargestDataset2019}. They usually consider 
many semantic categories (e.g., caption, paragraph, title) and split text regions at 
paragraph level. To perform text recognition, text line detection needs to be performed with 
dedicated methods such as described in the next paragraph. We argue that for many practical 
applications on historical documents in which layout is often simple, important elements are 
illustrations and text lines. In contrast to prior work, we thus target segmenting 
illustrations and text lines jointly, a problem we refer to as \textit{line-level page 
segmentation}.

\textbf{\textit{Text line detection.}}
While text line detection in modern printed documents is considered as a solved problem, it 
remains challenging for historical documents~\cite{simistiraICDAR2017CompetitionLayout2017,
diemCBADICDAR2017Competition2017, clausnerICFHR2018Competition2018, 
diemCBADICDAR2019Competition2019}. In most recent 
competitions~\cite{diemCBADICDAR2017Competition2017, diemCBADICDAR2019Competition2019} the 
task is actually to detect text baselines, which represent a compromise between annotation 
cost and descriptive power. We use instead the x-height 
representation~\cite{rentonHandwrittenTextLine2017} which not only enables to infer the 
baseline but also we believe to be more robust, easier to generalize and more directly useful 
for downstream text recognition tasks. Recent competitions were dominated by deep learning 
based approaches.The ICDAR2017 competition on BAseline Detection 
(cBAD2017)~\cite{diemCBADICDAR2017Competition2017} was won by the approach proposed by Fink 
et al.~\cite{finkBaselineDetectionHistorical2018}, a sliding-window dense prediction using a 
U-Net architecture~\cite{ronnebergerUnetConvolutionalNetworks2015}. Later, the winning entry 
was successively surpassed by the ResNet~\cite{heDeepResidualLearning2016} adaptation of Ares 
Oliveira et al.~\cite{aresoliveiraDhSegmentGenericDeepLearning2018} and by the model proposed 
by Grüning et al.~\cite{gruningTwostageMethodText2019} which added an attention mechanism and 
developed a sophisticated post-processing step based on superpixels. A slightly refined 
version of the latter also won the cBAD2019 
challenge~\cite{diemCBADICDAR2019Competition2019}. We use a plain segmentation approach 
similar to~\cite{aresoliveiraDhSegmentGenericDeepLearning2018} followed by a simple 
post-processing step designed to work for both text lines and illustrations.

\textbf{\textit{Synthetic data.}}
Training deep networks for both page and text line segmentation requires large amounts of 
data. For modern documents, Yang et al.~\cite{yangLearningExtractSemantic2017} and Zhong et 
al.~\cite{zhongPubLayNetLargestDataset2019} proposed synthetic document generation engines 
based on modern formats (respectively Latex and PDF) yielding to large-scale and 
heterogeneous document datasets. However, these documents are too simple to train a model 
that perform well on historical documents. To overcome the issue, Capobianco and 
Marinai~\cite{capobiancoDocEmulToolkitGenerate2017} as well as Journet et 
al.~\cite{journetDocCreatorNewSoftware2017} proposed toolkits to expand an existing annotated 
document dataset by generating similar semi-synthetic documents with the help of advanced 
data augmentation strategies. Resulting datasets are thus limited in diversity and they are 
not designed for generalization to new datasets. Besides, all the proposed generation 
processes either don't include graphical elements, or rely on very simple ones. On the 
contrary, we propose a complete synthetic document generation approach that generalizes well 
to a large variety of historical document datasets for both text line and illustration 
segmentation. Note that our synthetic documents can also be used for text recognition, 
similar to~\cite{guptaLearningReadSpelling2018}.

\section{Approach}

We consider line-level page segmentation as a pixel-wise classification task and propose to 
solve it using a deep neural network trained on our large-scale synthetic document dataset 
\textit{SynDoc}, followed by a simple connected component filtering. In this section, we 
first introduce our synthetic data generation engine, then describe our segmentation method.

\subsection{Synthetic document generation and labeling}\label{sec:syndoc}

\begin{figure*}
  \centering
  \includegraphics[height=3cm, width=0.495\textwidth]{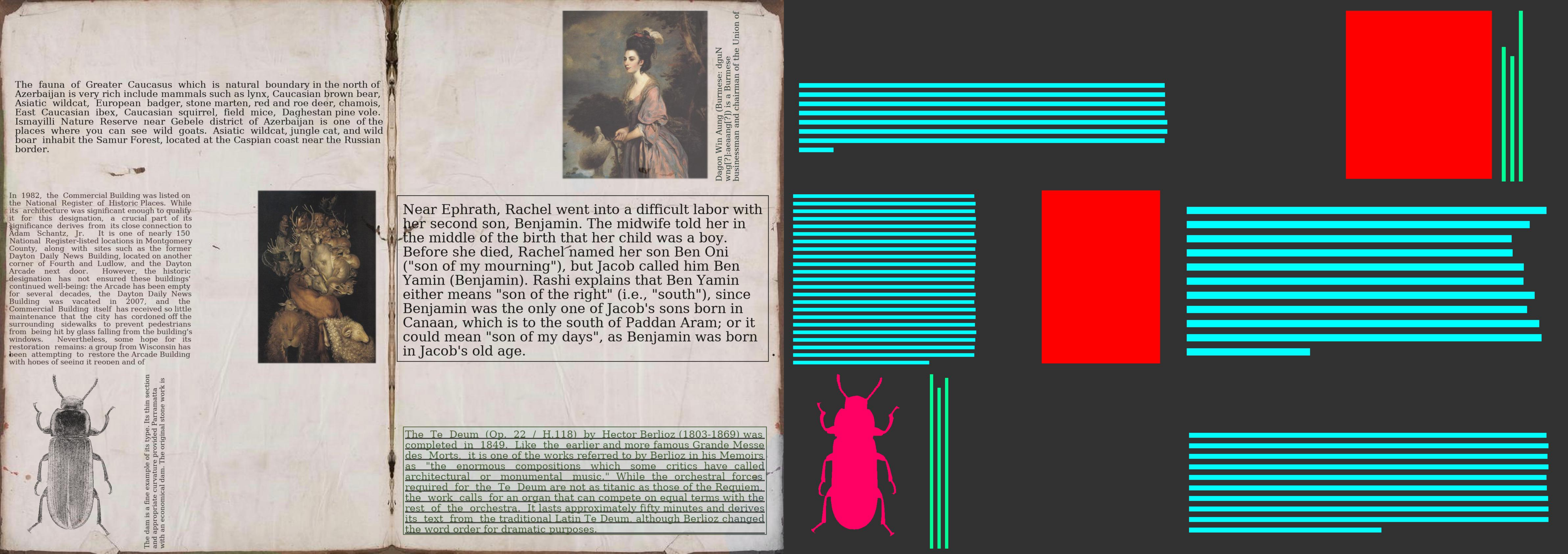}
  \includegraphics[height=3cm, width=0.2435\textwidth]{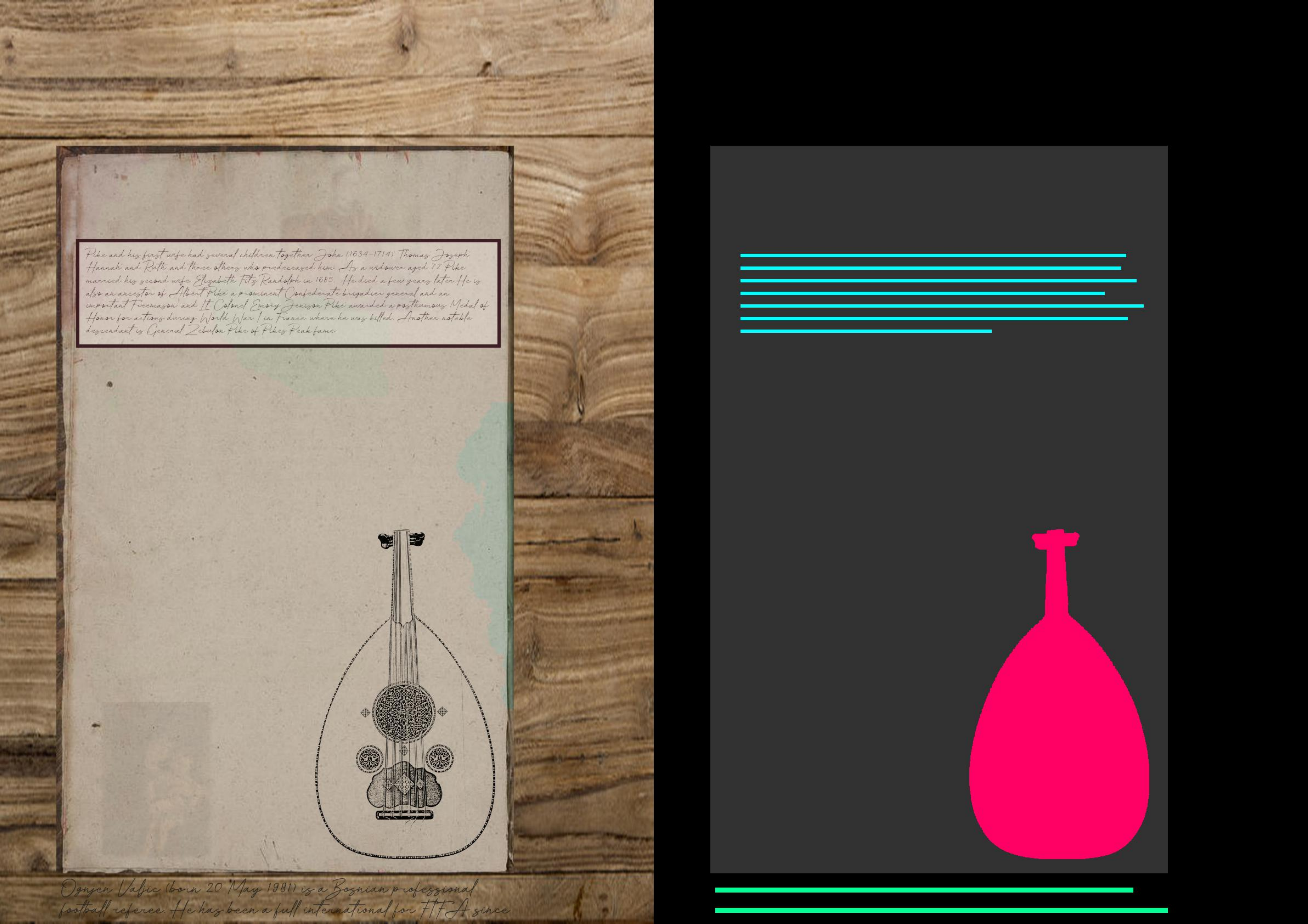}
  \includegraphics[height=3cm, width=0.2435\textwidth]{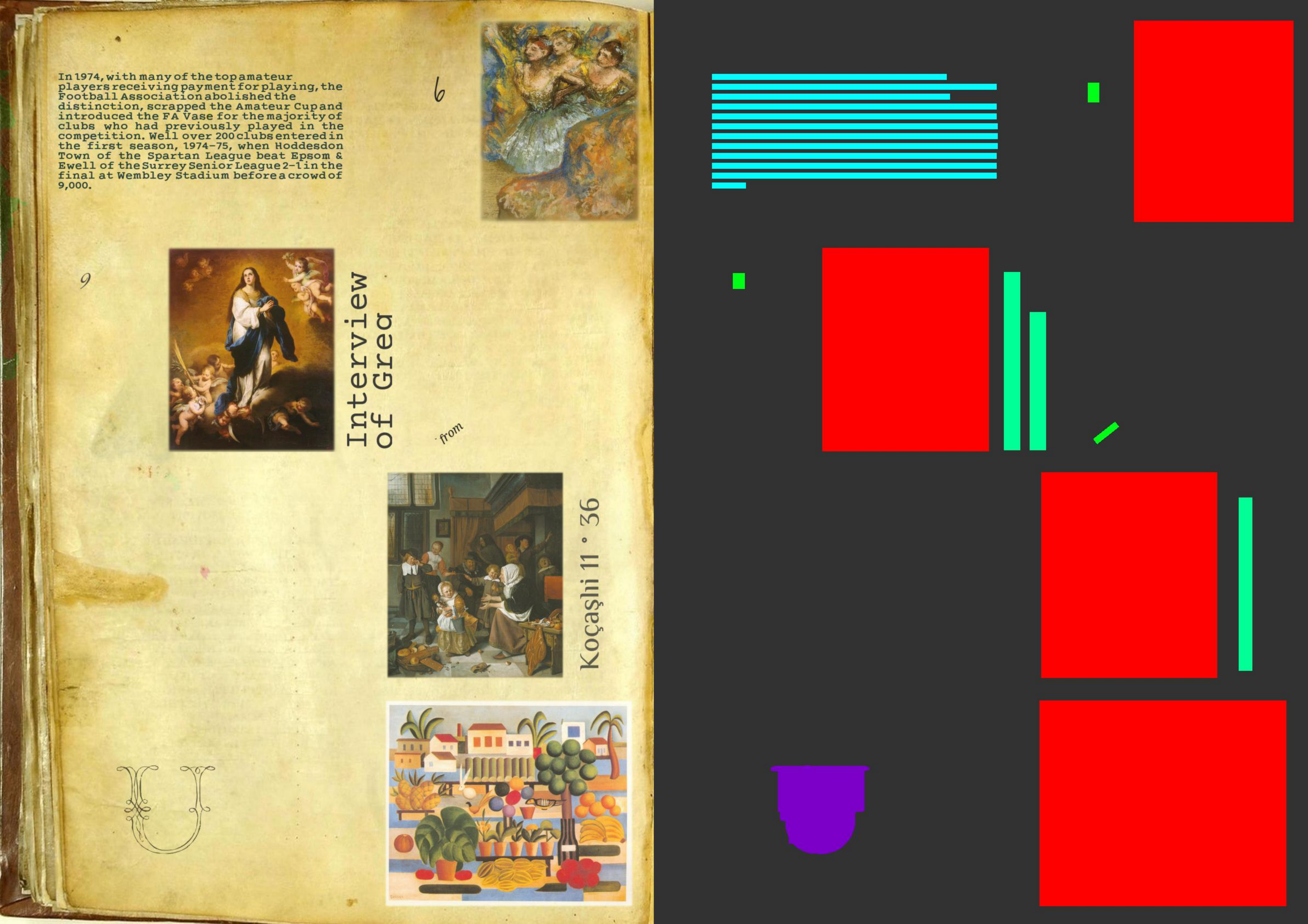}\\
  \vspace{0.3em}
  \includegraphics[height=3cm, width=0.2435\textwidth]{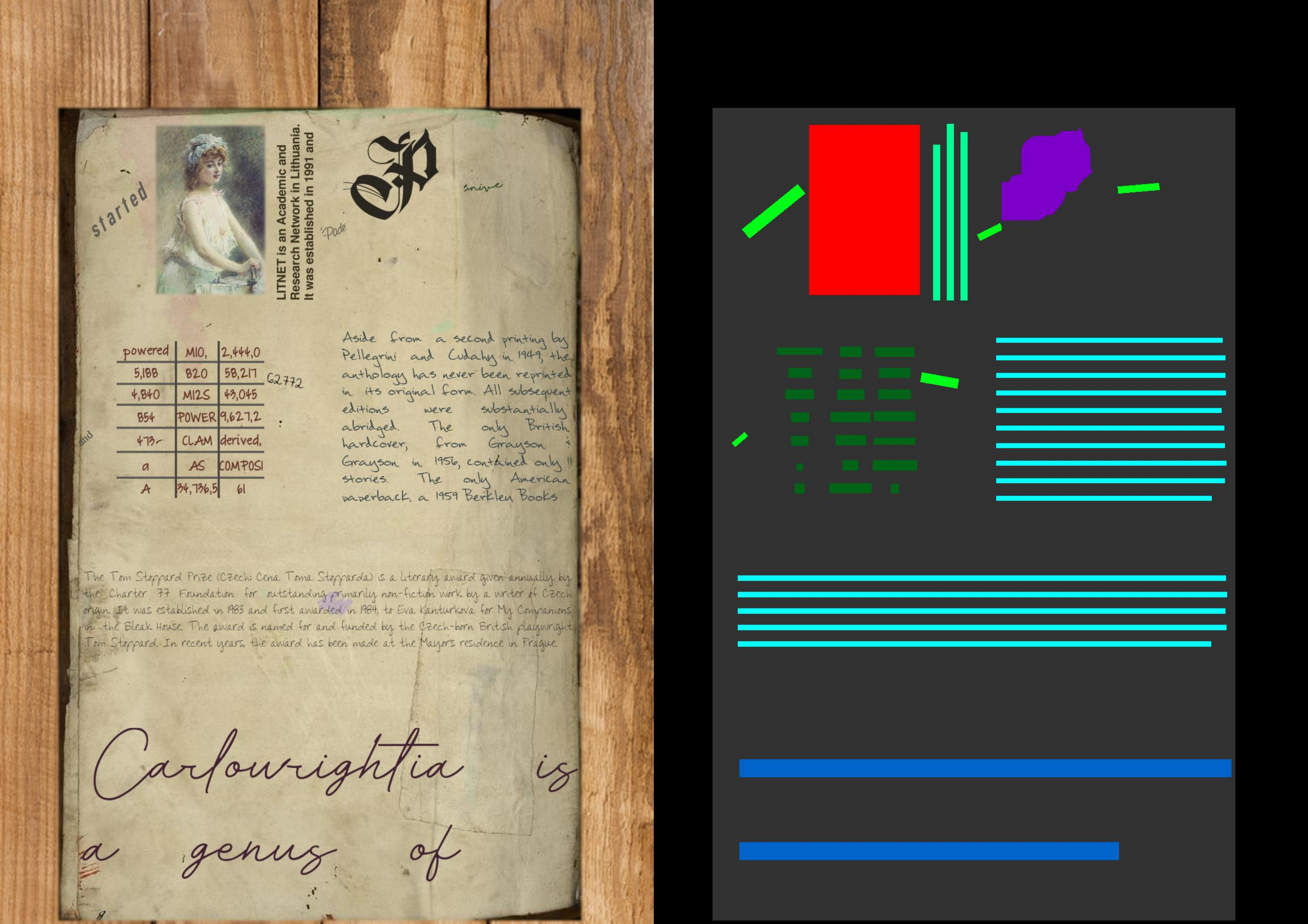}
  \includegraphics[height=3cm, width=0.2435\textwidth]{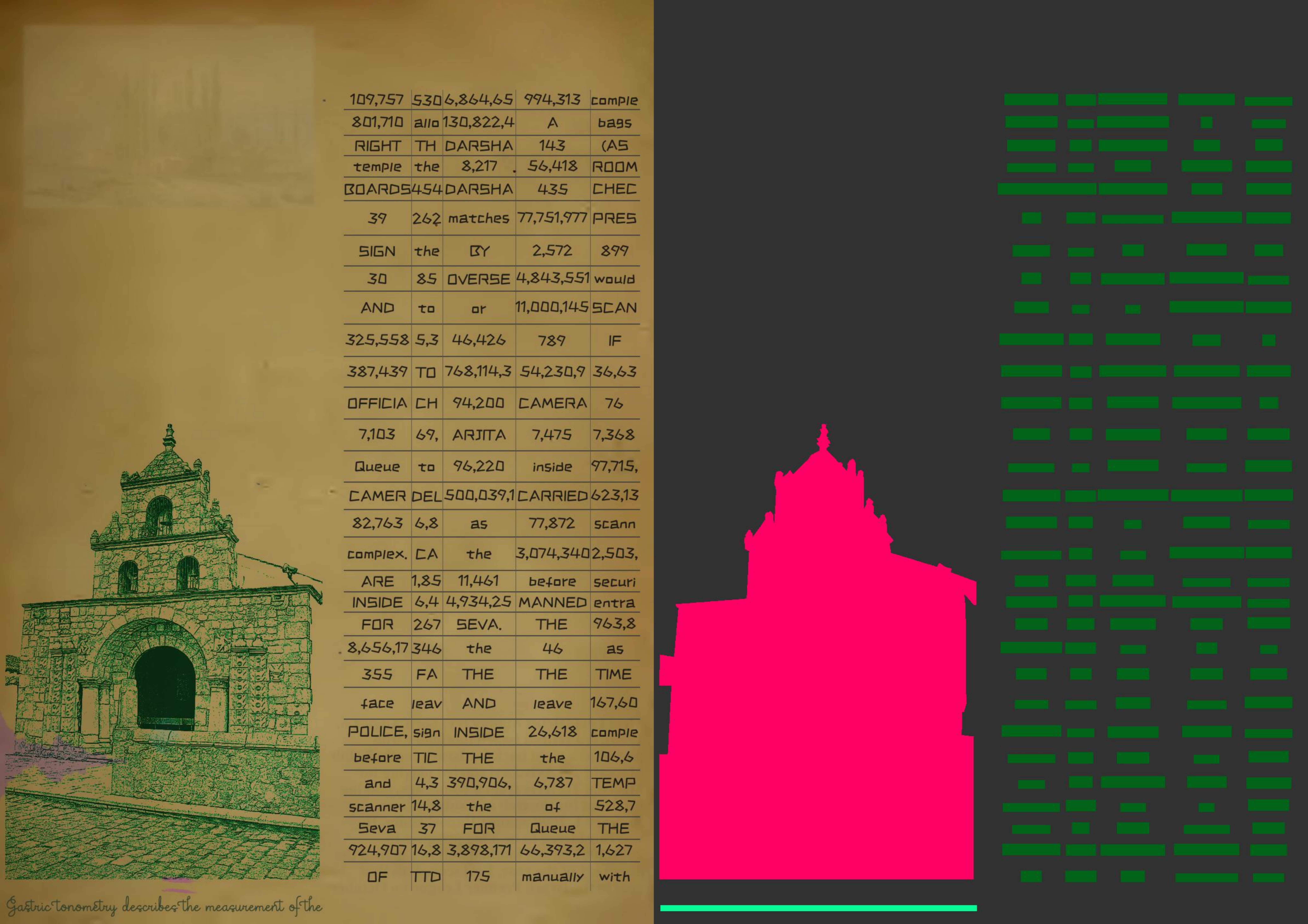}
  \includegraphics[height=3cm, width=0.495\textwidth]{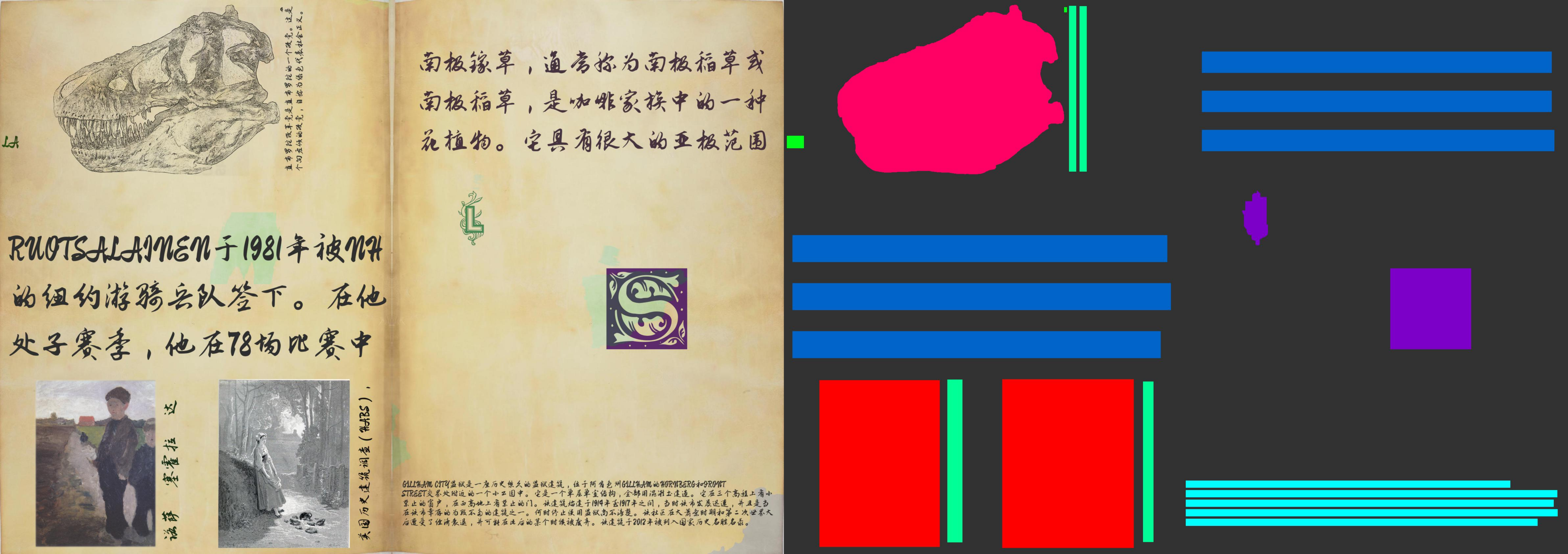}\\
  
  \caption{SynDoc examples with ground-truth. Elements are: 
    \colorbox{page-background}{\textcolor{white}{page-bkg}}, \colorbox{paragraph}{paragraph}, 
    \colorbox{table}{\textcolor{white}{table}}, \colorbox{title}{\textcolor{white}{title}},
    \colorbox{caption}{caption},
    \colorbox{floating-word}{floating-word},
  \colorbox{image}{\textcolor{white}{image}}, \colorbox{drawing}{\textcolor{white}{drawing}}, 
\colorbox{glyph}{\textcolor{white}{glyph}}.}
  \label{fig:syndoc}
\end{figure*}

While several datasets~\cite{antonacopoulosRealisticDatasetPerformance2009, 
clausnerENPImageGround2015, yangLearningExtractSemantic2017, 
zhongPubLayNetLargestDataset2019} are available for page segmentation, they do not embrace 
the wide diversity of historical documents. Furthermore, text regions are always annotated as 
coarse text blocks preventing straightforward line extractions. To address these issues, we 
created a fast and scalable synthetic document generation engine with pixel-wise annotations 
and use it to generate a dataset of 10k images called \textit{SynDoc}. We first present an 
overview of the generation process, then the basic elements we used to obtain challenging 
data and finally the labeling we designed for optimal generalization. Examples of generated 
documents can be seen in Fig.~\ref{fig:syndoc}.

\subsubsection{\textbf{Document generation process}}\label{sec:syndoc_gen}
The document generation process includes three randomized steps. First, a page background is 
selected from a set of 177 empty pages we collected and undergoes augmentations: it can be 
symmetrized to mimic a double page or pasted on a contextual image picked from a set of 15 
images. Second, a grid page layout is drawn and each empty cell is filled with an element 
with random margins. In the case the element is graphical, a horizontal and vertical caption 
can be added. Third, different forms of degradations are applied to avoid overfitting and 
increase robustness: Gaussian blur, structured noise (random shapes) addition and 
bleed-through. Bleed-through degradation is critical in manuscript layout analysis and we 
perform it by overlaying another grid of random elements with low opacity. This modular 
approach enables to easily add new types of elements.

\subsubsection{\textbf{Element generation}}\label{sec:element_gen}
We implemented four types of elements which we found to be critical to obtain good results on 
real historical documents:

\begin{itemize}
  \item \textit{text}: we used texts scraped from random wikipedia pages. Texts can then be 
    generated in 5 different layouts: \textit{caption}, \textit{floating-word}, 
    \textit{paragraph}, \textit{table} and \textit{title}. We augment them using translation 
    (to Arabic or Chinese), font changes (selection from 405 fonts we downloaded from the 
    web\footnotemark[2] and formatting such as size or spacing), justification, strike 
    through, underlining, rotation and bounding box addition,
  \item \textit{image}: we used the Wikiart dataset\footnote{http://www.wikiart.org/} which 
    contains much more difficult images for page segmentation than natural image datasets.
  \item \textit{drawing}: we transformed images scraped from random wikipedia pages into 
    drawings by blending them with their blurred negative through color dodging,
  \item \textit{glyph}: we collected 91 decorated fonts from the 
    web\footnote{\label{fn:font}https://www.dafont.com/, https://fonts.google.com/} and a 
    random uppercase letter is picked to generate a glyph.
\end{itemize}

At generation time, we perform generic on-the-fly augmentations such as blurring, 
colorization and opacity variation. Each element class is associated to its own labeling 
described in the next paragraph. The benefits of using background augmentations, drawings, 
glyphs, text translation and bleed-through are experimentally demonstrated in 
Sec.~\ref{sec:ablation}.

\subsubsection{\textbf{Element labeling}}\label{sec:labeling}
Even though labeling can be element-specific, we argue a wide diversity of labels makes it 
difficult to generalize to new types of documents. Following most text line detection 
competitions~\cite{diemCBADICDAR2017Competition2017, clausnerICFHR2018Competition2018, 
diemCBADICDAR2019Competition2019}, we thus label all text elements the same way and associate 
to all graphical elements a single \textit{illustration} label. 

While labeling images is straightforward, we label the shape of glyphs and drawings using 
closing operations. Contrary to bounding box or contour labels, we believe such labeling is 
not only accurate enough to extract the targeted region without surrounding elements but also 
coarse enough to be easily learned by the model. For text, to perform page segmentation at 
line level, we adopted x-height representation, which corresponds to the core area of the 
text without ascenders and descenders. Unlike bounding box or baseline labels, it enables a 
straightforward line extraction while preventing lines from merging. Besides, we expect that 
x-height representation has a better generalization power to unknown fonts than baseline 
labels as it doesn't require to infer text orientation. Because we still observed lines 
vertically merged by small pixel bridges in the case of thin interline spaces, we labeled 
border regions around the text representations to help the model learn interline spaces. We 
experimentally show in Sec.~\ref{sec:ablation} the improvements stemming from such text 
labeling choices.

\subsection{Segmentation method}\label{sec:method}
We perform line-level page segmentation using a fully convolutional network, optimized with a 
standard cross-entropy loss and followed by a simple post-processing.

\begin{figure}[bp]
    \centering
    \includegraphics[width=\columnwidth]{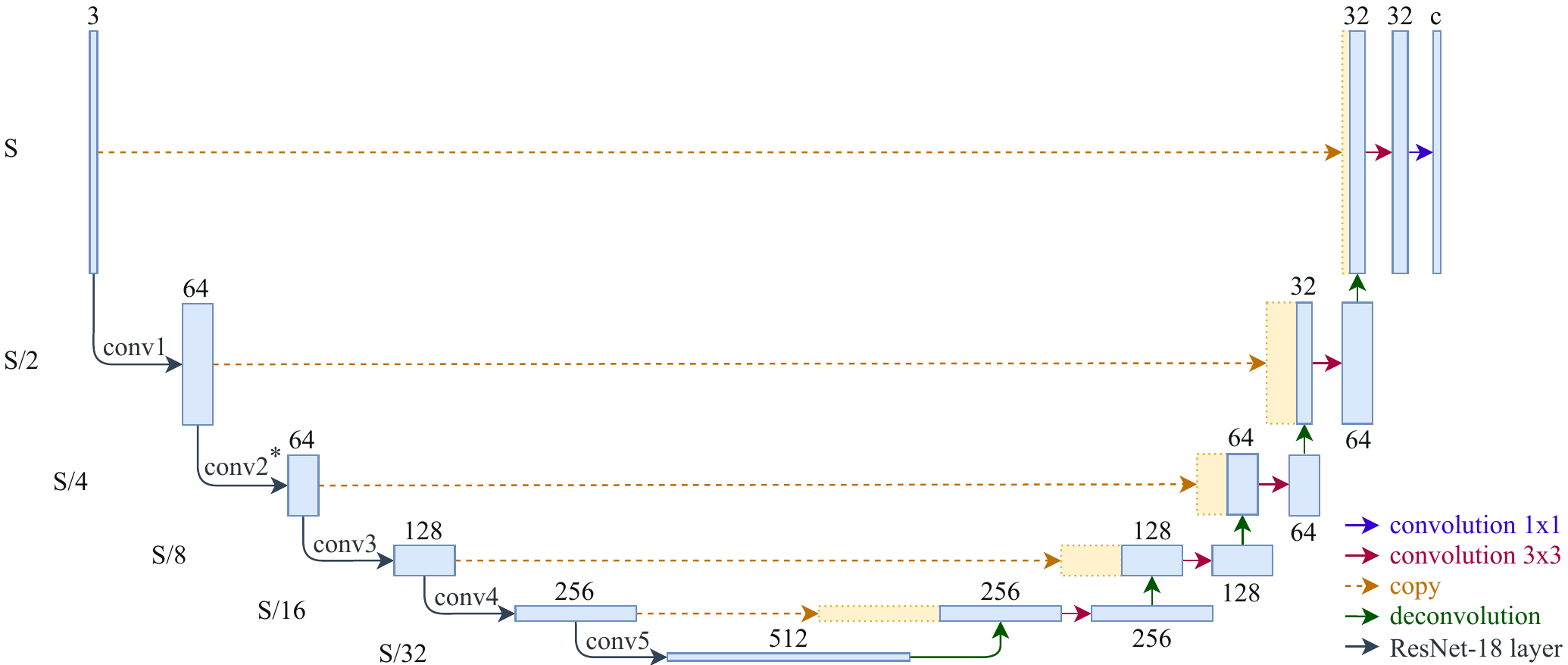}
    \caption{Network architecture. \textsuperscript{*} indicates max-pooling replacement.}
    \label{fig:architecture}
\end{figure}

\subsubsection{\textbf{Network architecture}}\label{sec:architecture}
Similar to~\cite{aresoliveiraDhSegmentGenericDeepLearning2018}, we use a simple 
encoder-decoder architecture combining the descriptive power of 
ResNet~\cite{heDeepResidualLearning2016} with the localization recovering capacity of 
U-Net~\cite{ronnebergerUnetConvolutionalNetworks2015}. Compared 
to~\cite{aresoliveiraDhSegmentGenericDeepLearning2018}, we use a smaller ResNet-18 as 
backbone encoder since detecting text lines requires keeping document images as large as 
possible, which constraints memory, and we perform small modifications in the architecture 
resulting in better performances. The full network architecture is summarized in 
Fig.~\ref{fig:architecture}.

We replaced the max-pooling operation in ResNet conv2 block by a 2-strided 3x3 convolutional 
layer, as max-pooling has been shown~\cite{yuDilatedResidualNetworks2017} to lead to gridding 
artifacts. The decoder is composed of 5 upscaling blocks and a final convolutional layer 
which assigns a class to each pixel. Each upscaling block is composed of an upscaled version 
of the previous feature map concatenated with the corresponding encoding feature map and a 
3x3 convolutional layer. Because text lines can be small, we upsample features using 
deconvolutional~\cite{nohLearningDeconvolutionNetwork2015} layers with stride 2 rather than 
bilinear interpolation.

\subsubsection{\textbf{Post-processing}}
We use a simple post-processing step filtering out connected components with low area using a 
class-specific ratio threshold.

To compare with state-of-the-art baseline detection methods, we either retrieve baselines 
from the segmentation maps, or for low shot comparisons change our labels to directly predict 
baselines. To compute a baseline from a x-height component, we first fit a straight line to 
retrieve the text orientation and its bottom line. We then fit to the latter a 5-degree 
polynomial to get a smooth baseline prediction. This process assumes that the page is well 
oriented and it particularly fails in the case of transposed texts (90\si{\degree} or 
180\si{\degree} rotated).

\subsubsection{\textbf{Implementation details}}\label{sec:train_details}

All images are resized so that their larger side is 1280 pixels, keeping the aspect ratio 
constant. We perform per-channel standardization and during training several on-the-fly data 
augmentations including Gaussian blur, brightness and contrast variation, image rotation and 
transposition. When fine-tuning on real datasets, we also perform random scaling which we 
found to be critical for high-performances. We limit the maximum number of pixels to 
$3.5\times10^{6}$ to avoid memory error while scaling.
For memory reasons, we process one sample per batch and use Instance 
Normalization~\cite{ulyanovInstanceNormalizationMissing2016} with a momentum of 0.1 instead 
of batch normalization. We use ImageNet~\cite{dengImageNetLargescaleHierarchical2009} 
pre-trained weights for the encoder, which significantly speeds up the training, and Xavier 
initialization~\cite{glorotUnderstandingDifficultyTraining2010} for the other convolutional 
layers. We train for 100 epochs with Adam optimizer~\cite{kingmaAdamMethodStochastic2015} 
with a weight decay of $10^{-6}$. Learning rate is initially set to 0.001 and divided by 2 
after 30, 60 and 80 epochs. On a Nvidia GeForce RTX 2080 Ti GPU, training takes approximately 
3 days and single image inference takes 1.06 second.

\section{Experiments}

In this section, we first introduce the sets of datasets used for text line detection and 
illustration segmentation evaluations, including our new \textit{IlluHisDoc}. Then, we 
present quantitative results with comparisons to state-of-the-art of both our off-the-shelf 
and fine-tuned approach. Finally, we present a methodical ablation study of our approach.

\subsection{Datasets}

The cBAD competitions~\cite{diemCBADICDAR2017Competition2017, 
diemCBADICDAR2019Competition2019} involve large datasets with a great variety of historical 
document images and are the standard benchmarks for text line detection. To the best of our 
knowledge, there is no dataset for illustration segmentation with such a diversity. Hence, we 
evaluate our method using three diverse datasets: 
Mandragore\footnote{http://api.bnf.fr/mandragore-echantillon-segmente-2019},
RASM2019\footnote{https://www.primaresearch.org/RASM2019/resources} and our proposed 
IlluHisDoc dataset.

\subsubsection{\textbf{cBAD2017 and cBAD2019}}
Dataset for cBAD2017 is split in two, Simple and Complex Tracks, with respectively 216 and 
270 images for training, 539 and 1010 images for evaluation. Larger and more diversified, 
cBAD2019 contains 1510 training and 1511 evaluation images.

\subsubsection{\textbf{Mandragore}}
Dedicated to the illustration detection, the dataset is composed of 8 manuscripts, gathering 
2807 pages including 631 illustrations annotated with bounding boxes. Because of inconsistent 
annotations, we removed \textit{Français 2692} and \textit{Latin 757} manuscripts as well as 
all book spine and cover images, resulting in a dataset of 1691 images.

\subsubsection{\textbf{RASM2019}}
Dataset is composed of Arabic scientific handwritten manuscripts. Initially meant for text 
detection and recognition, it also includes annotations for scientific figures labeled as 
graphics and images, which we merged into a global illustration class. Test set consists in 
100 images.

\begin{figure*}
  \centering
  \includegraphics[height=3cm, width=0.32\columnwidth]{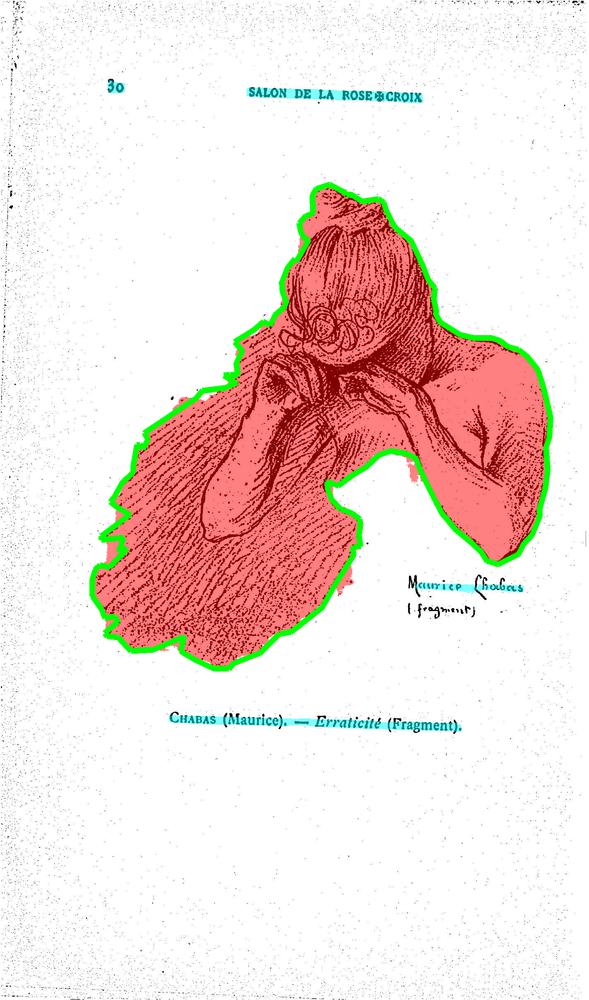}
  \includegraphics[height=3cm, width=0.64\columnwidth]{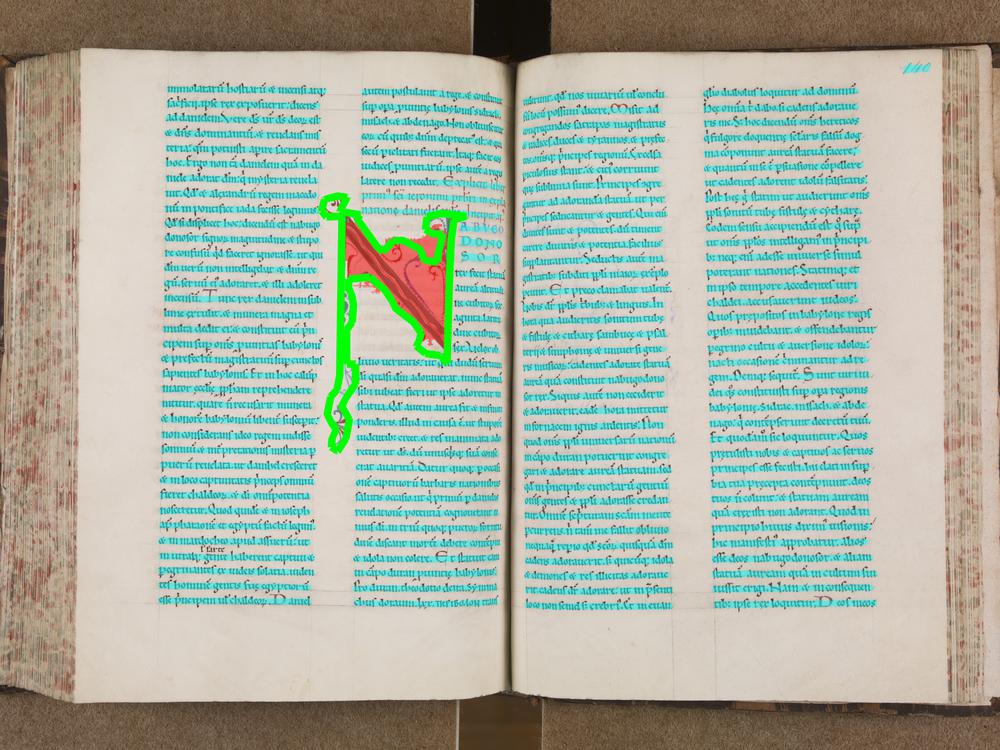}
  \includegraphics[height=3cm, width=0.64\columnwidth]{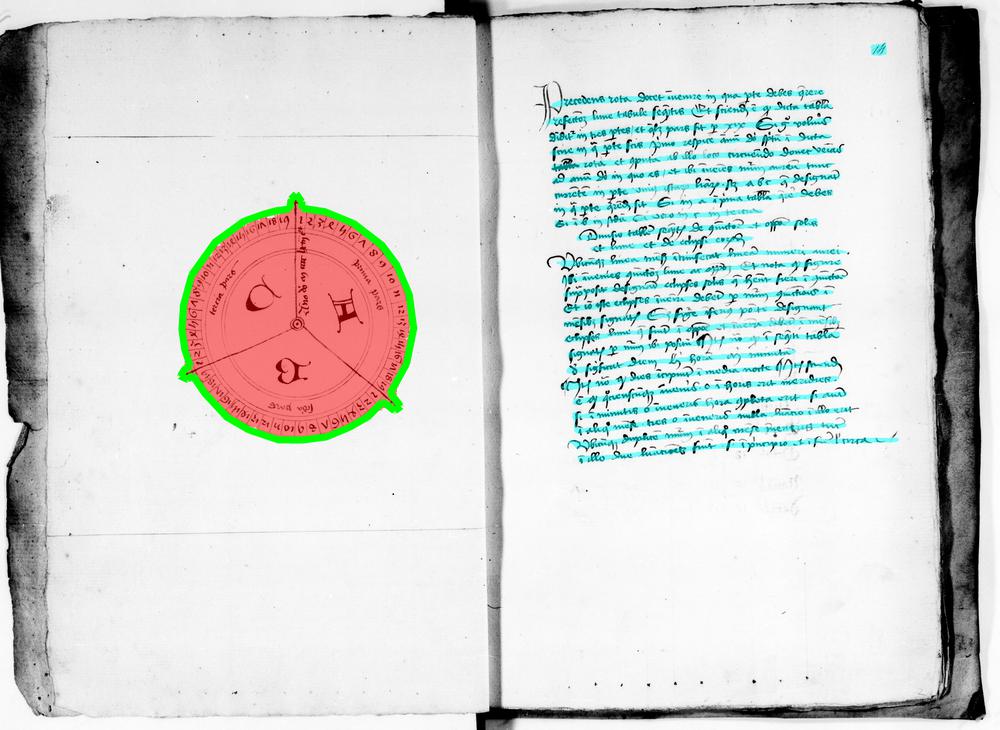}
  \includegraphics[height=3cm, width=0.32\columnwidth]{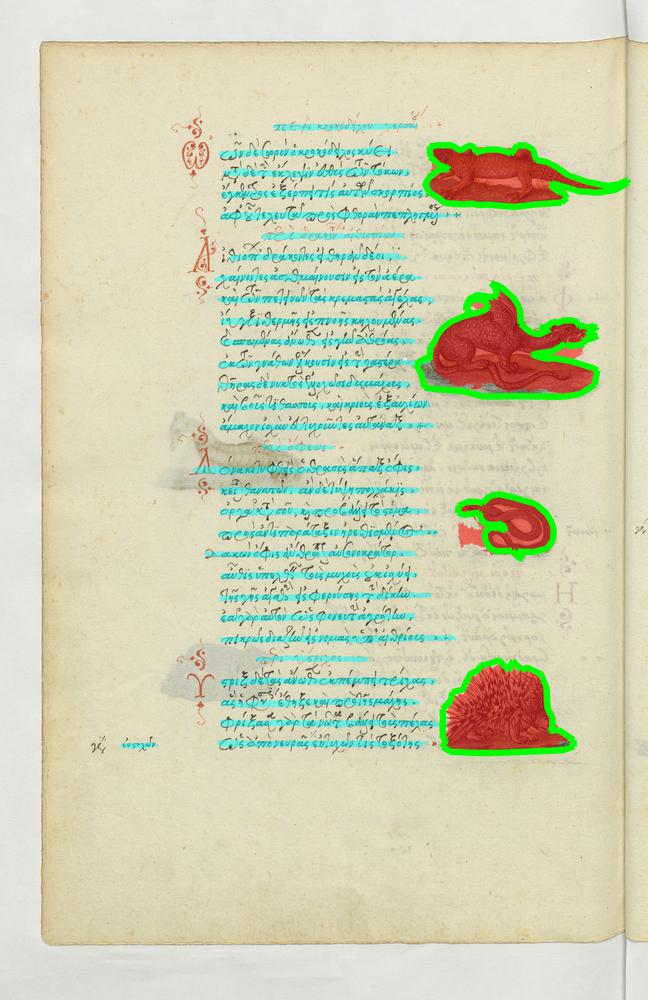}
  
  \caption{Examples from IlluHisDoc dataset with \textcolor{gt}{ground-truth} and 
    segmentation outputs of our method (\colorbox{illustration}{illustration} and
    \colorbox{text}{text}). From left to right: printed documents (P), manuscripts with 
  illuminations (MSI), manuscripts with scientific diagrams (MSS), manuscripts with drawings 
(MSD).}
    \label{fig:illuhisdoc}
\end{figure*}

\subsubsection{\textbf{IlluHisDoc}}
To provide a more representative evaluation for illustration segmentation, we created a new 
test dataset dubbed IlluHisDoc (Illustrated Historical Documents). We designed it to include 
diverse types of illustrations relevant for digital humanities and to embrace a wide variety 
of documents, layouts and degradations. Document images were mainly downloaded from 
Gallica\footnote{\label{fn:gallica}gallica.bnf.fr, Bibliothèque nationale de France}. We 
explicitly split IlluHisDoc in 4 parts corresponding to different types of illustrations:

\begin{itemize}
  \item \textit{P}: 5 printed documents that comprise multiple forms of illustration 
    (drawing, ornament, painting, photo),
  \item \textit{MSS}: 5 manuscripts with scientific diagrams,
  \item \textit{MSI}: 5 manuscripts with illuminations,
  \item \textit{MSD}: 5 manuscripts with drawings.
\end{itemize}

In each source document, we annotated 10 images with at least one illustration and 10 images 
without any resulting in 400 pages. Annotations were performed at pixel-level using VGG Image 
Annotator~\cite{duttaAnnotationSoftwareImages2019}.
Note that the aim of this dataset is to evaluate the generalization capability of 
out-of-the-box solutions to generalize to unseen data and not for training. Examples of our 
four types of documents are shown in Fig.~\ref{fig:illuhisdoc}.

\subsection{Results}

In this section, we evaluate our approach for both baseline detection and illustration 
segmentation. We compare to the off-the-shelf open-source Tesseract4 and state-of-the-art 
methods.

\begin{table}
  \caption{Results for cBAD2017 dataset}
  \addtolength{\tabcolsep}{-3pt}
  \centering
  \vspace{-0.4em}
  \begin{tabular}{@{}lccccccc@{}} \toprule
  Method & Training & \multicolumn{3}{c}{Simple Track} & \multicolumn{3}{c}{Complex Track}\\
  \cmidrule(lr){3-5} \cmidrule(l){6-8} & set used & P-val & R-val & F-val & P-val & R-val & 
  F-val\\
  \midrule
  Tesseract4 && 0.396 & 0.545 & 0.459 & 0.322 & 0.520 & 0.398\\
  \textbf{Ours (off-the-shelf)} & & \textbf{0.871} & \textbf{0.930} & \textbf{0.900} & 
  \textbf{0.844} & \textbf{0.782} & \textbf{0.812}\\
  \midrule
  LITIS~\cite{rentonHandwrittenTextLine2017,diemCBADICDAR2017Competition2017} & \checkmark & 
  0.780 & 0.836 & 0.807 & - & - & -\\
  IRISA~\cite{diemCBADICDAR2017Competition2017} & \checkmark & 0.883 & 0.877 & 0.880 & 0.692 
  & 0.772 & 0.730\\
  UPVLC~\cite{diemCBADICDAR2017Competition2017} & \checkmark & 0.937 & 0.855 & 0.894 & 0.833 
  & 0.606 & 0.702\\
  BYU~\cite{diemCBADICDAR2017Competition2017} & \checkmark & 0.878 & 0.907 & 0.892 & 0.773 & 
  0.820 & 0.796\\
  dhSegment~\cite{aresoliveiraDhSegmentGenericDeepLearning2018} & \checkmark & 0.88 & 0.97 & 
  0.92 & 0.79 &\bf 0.95 & 0.86\\
  DMRZ~\cite{finkBaselineDetectionHistorical2018,diemCBADICDAR2017Competition2017} & 
  \checkmark & 0.973 & 0.970 & 0.971 & 0.854 & 0.863 & 0.859\\
  Planet~\cite{gruningTwostageMethodText2019} & \checkmark & \textbf{0.98} & \bf 0.98 & 
  \textbf{0.978} & \textbf{0.93} & 0.92 & \textbf{0.922}\\
  \bf Ours (fine-tuned)& \checkmark & 0.948 & \textbf{0.978} & 0.963 & 0.883 & \textbf{0.947} 
  & 0.914\\
  \bottomrule
  \end{tabular}
  \label{tab:bad2017}
  \vspace{-0.4em}
\end{table}

\begin{table}
  \caption{Results for cBAD2019 dataset}
  \centering
  \vspace{-0.4em}
  \begin{tabular}{@{}lcccc@{}}
  \toprule Method & Training set used & P-val & R-val & F-val\\
  \midrule
  Tesseract4 & & 0.442 & 0.552 & 0.491\\
  \bf Ours (off-the-shelf) & & \textbf{0.844} & \textbf{0.815} & \textbf{0.829}\\
  \midrule
  Baseline~\cite{diemCBADICDAR2019Competition2019}& \checkmark & 0.773 & 0.743 & 0.758\\
  TJNU~\cite{diemCBADICDAR2019Competition2019} & \checkmark & 0.852 & 0.885 & 0.868\\
  UPVLC~\cite{diemCBADICDAR2019Competition2019} & \checkmark & 0.911 & 0.902 & 0.907\\
  DMRZ~\cite{diemCBADICDAR2019Competition2019} & \checkmark & 0.925 & 0.905 & 0.915\\
  Planet~\cite{diemCBADICDAR2019Competition2019} & \checkmark & \textbf{0.937} & 0.926 & 
  \textbf{0.931}\\
  \bf Ours (fine-tuned)& \checkmark & 0.920 & \textbf{0.931} & 0.925\\
  \bottomrule
  \end{tabular}
  \label{tab:bad2019}
  \vspace{-0.4em}
\end{table}

\subsubsection{\textbf{Baseline detection}}
We evaluate our method for baseline detection on the test split of cBAD2017 and cBAD2019 
using the competition evaluation scheme. We report results for off-the-shelf and fine-tuned 
configurations in Table~\ref{tab:bad2017} and~\ref{tab:bad2019}.

While never trained on real data, our off-the-shelf approach provides good results
across the three benchmarks, outperforming Tesseract4 by a large margin and showing results 
comparable to the weaker methods trained on real data. This shows that our method trained 
only on computer-generated data generalizes well to real and complex handwritten data.

After fine-tuning on the respective training sets, our method leads to results on par with 
state-of-the-art methods. This is a strong result since these methods typically involve very 
advanced and specific post-processing steps, while ours is very simple and common across all 
datasets and element types. Our performance could likely be further improved, for example 
using the superpixel-based post-processing of~\cite{gruningTwostageMethodText2019} which they 
demonstrate to provide a remarkable boost. This is however orthogonal to the goal of our 
work, which is providing a simple, robust and generic method. 

\begin{figure}
    \centering
    \includegraphics[width=0.9\columnwidth]{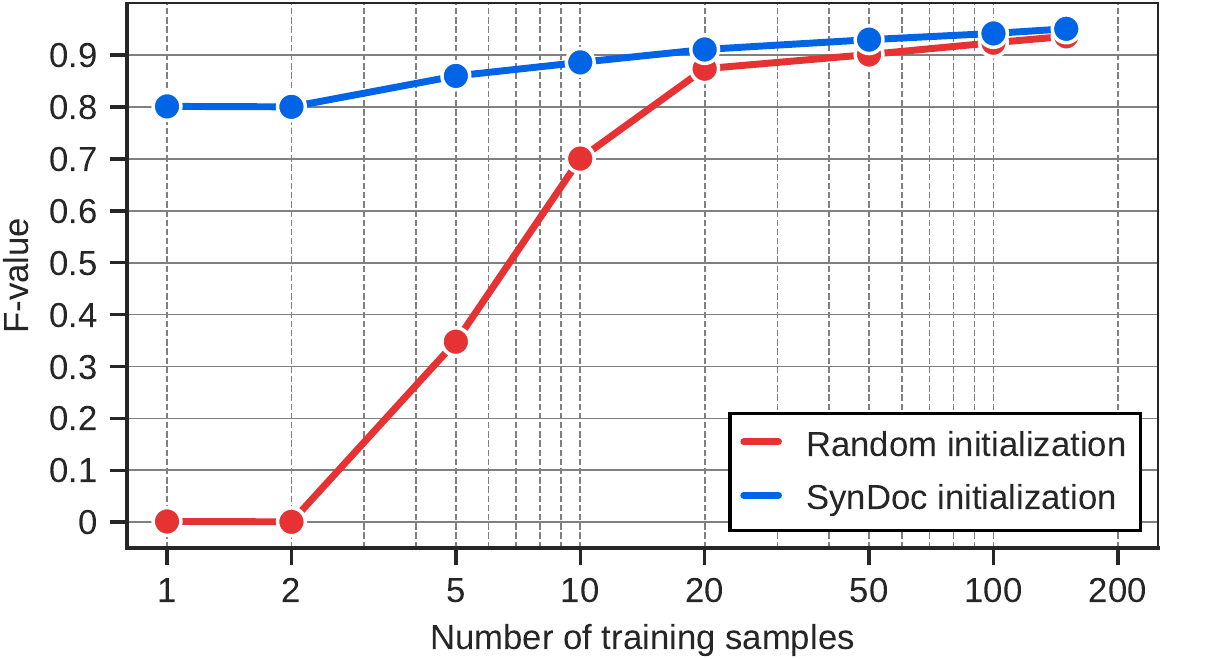}
    \caption{F-value with different amount of training data on cBAD2017 Simple Track using 
    networks randomly initialized and pre-trained on SynDoc.}
    \label{fig:finetunecbad2017}
\end{figure}

Our synthetic training data could also be used to initialize a network and fine-tune it with
a few real examples. To evaluate this setup, we split cBAD2017 Simple Track training set in 
two, keeping 40 samples as evaluation set and training on the rest. In 
Fig.~\ref{fig:finetunecbad2017}, we compare random and SynDoc initializations with an 
increasing number of training samples to fine-tune the network. Remarkably, SynDoc 
initialization consistently leads to better results, with particularly large gaps when less 
than 20 annotated samples are available.

\subsubsection{\textbf{Illustration segmentation}}
We evaluate our off-the-shelf approach for illustration segmentation on Mandragore, RASM2019 
and IlluHisDoc. Mean Intersection over Union (mIoU) scores are reported in 
Table~\ref{tab:illu}.
 
To validate the benefits of both our synthetic dataset and our segmentation approach, we
also report results using the synthetic PubLayNet dataset introduced 
in~\cite{zhongPubLayNetLargestDataset2019} as well as one of their benchmarked methods, 
Mask-RCNN~\cite{heMaskRCNN2017}. Our full method provides results far above any baseline, 
including Tesseract4 which generalizes poorly to historical documents. Two effects can be 
clearly identified. First, for both training datasets, our segmentation approach provides 
much better results than a detection-based approach across all datasets. Second, training on 
SynDoc provides much better generalization on historical datasets than training on PubLayNet 
both for Mask-RCNN and our segmentation method. 

\begin{table}[t]
  \caption{Illustration segmentation (mIoU in \%)}
  \addtolength{\tabcolsep}{-2pt}
  \centering
  \vspace{-0.4em}
  \begin{tabular}{@{}llccccccc@{}} \toprule
  &&&& \multicolumn{5}{c}{IlluHisDoc}\\
  \cmidrule(l){5-9} Method & Training & Mandra. & RASM & avg & P & MSS & MSI & MSD\\
  \midrule
  Tesseract4 && 17.2 & 6.0 & 14.8 & 41.4 & 9.2 & 2.0 & 6.5\\
  M-RCNN & PubLay. & 9.8 & 4.2 & 11.5 & 34.3 & 3.1 & 1.5 & 7.2\\
  Ours & PubLay. & 18.3 & 14.8 & 24.3 & 57.7 & 16.9 & 5.3 & 17.2\\
  M-RCNN & SynDoc & 72.3 & 36.9 & 55.4 & 93.5 & 60.3 & 40.3 & 27.6\\
  \bf Ours &\bf SynDoc & \textbf{86.6} & \textbf{71.0} & \textbf{76.1} & \textbf{97.2} & 
  \textbf{61.8} & \textbf{76.8} & \textbf{68.5}\\
  \bottomrule
  \end{tabular}
  \label{tab:illu}
  \vspace{-0.4em}
\end{table}

\subsection{Ablations}\label{sec:ablation}

We here show the benefits of our contributions for synthetic document generation, text 
labeling and network architecture. All experiments are trained on SynDoc following
Sec.~\ref{sec:train_details}.

\begin{table}[t]
  \caption{Ablation study on SynDoc improvements (mIoU in \%)}
  \addtolength{\tabcolsep}{-1pt}
  \centering
  \vspace{-0.4em}
  \begin{tabular}{@{}lccc@{}}
  \toprule Experiment & Mandragore & RASM2019 & IlluHisDoc\\
  \midrule
  SynDoc & \textbf{86.6} & \textbf{71.0} & 76.1\\
  \quad w/o bleed-through & 84.3 & 67.0 & \textbf{77.2}\\
  \quad w/o text translation & 84.4 & 63.6 & 76.6\\
  \quad w/o drawing \& glyph & 80.5 & 23.6 & 52.5\\
  \quad w/o bkg augmentations & 55.9 & 44.6 & 44.1\\
  \bottomrule
  \end{tabular}
  \label{tab:syndoc}
  \vspace{-0.4em}
\end{table}

\subsubsection{\textbf{SynDoc}}
In Table~\ref{tab:syndoc}, we evaluate the improvements proposed for synthetic document 
generation by systematically removing them from the generation engine. Evaluation is done 
for illustration segmentation in Mandragore, RASM2019 and IlluHisDoc. Results show that 
adding bleed-through, texts in different languages, drawings and glyphs as well as augmenting 
page backgrounds with double pages or contextual images, all contribute to our high 
performances in amounts that differ depending on the specificity of each test dataset.

\begin{table}
  \caption{Ablation experiments for text labeling choices (F-value)}
  \centering
  \vspace{-0.4em}
  \begin{tabular}{@{}ccccc@{}}
  \toprule && \multicolumn{2}{c}{cBAD2017}&\\
  \cmidrule(lr){3-4}
  text label & border label & Simple & Complex & cBAD2019\\
  \midrule
  baseline & & 0.663 & 0.719 & 0.637\\
  baseline & \checkmark & 0.714 & 0.771 & 0.678\\
  x-height & & 0.749 & 0.724 & 0.758\\
  x-height & \checkmark & \textbf{0.900} & \textbf{0.812} & \textbf{0.829}\\
  \bottomrule
  \end{tabular}
  \label{tab:text_label}
  \vspace{-0.4em}
\end{table}

\subsubsection{\textbf{Text labeling}}
In Table~\ref{tab:text_label}, we show the benefits of our x-height representation with 
border labels for text lines. We train our approach with different labels and evaluate 
baseline detection on cBAD2017 and cBAD2019. Two main effects can be seen. First, predicting 
x-height representation and using the prediction to infer baselines performs better than 
directly predicting the baseline. Second, adding border labels dramatically boosts 
performances both when training with x-height and baseline representations. Nonetheless, the 
boost is much clearer when using x-height, because in this case borders are necessary to 
avoid merging different close lines. On the three benchmarks, the combination of x-height 
with border labels provides a very significant boost, allowing our method to perform well 
without advanced post-processing.

\begin{table}[t]
  \caption{Ablation experiments for architecture modifications evaluated on SynDoc (IoU in 
  \%)}
  \addtolength{\tabcolsep}{-2pt}
  \centering
  \vspace{-0.4em}
  \begin{tabular}{@{}cccccccc@{}}
  \toprule conv2 & upscaling & \#param & bkg & illustration & text & border & avg\\
  \midrule
  max-pooling & bilinear & 13.6M & 97.5 & 94.6 & 85.8 & 74.6 & 88.1\\
  strided conv & bilinear & 13.6M & 97.6 & \textbf{95.0} & 86.6 & 75.5 & 88.7\\
  strided conv & deconv & 14.4M & \textbf{97.7} & 94.9 & \textbf{87.6} & \textbf{77.2} & 
  \textbf{89.3}\\
  \bottomrule
  \end{tabular}
  \label{tab:archi}
  \vspace{-0.4em}
\end{table}

\subsubsection{\textbf{Network architecture}}
We now validate the benefits of the architecture changes we made compared to 
dhSegment~\cite{aresoliveiraDhSegmentGenericDeepLearning2018}: a simple ResNet-18 backbone, 
the replacement of the max-pooling by a strided convolution and the deconvolutional 
upscaling. In Table~\ref{tab:archi}, we evaluate three variants of our model on a synthetic 
testing set, with and without max-pooling and upscaling replacements using IoU for all 
labels. This enables us to obtain results similar to those of dhSegment on the same data 
(88.8\% in average compared to 89.3\% for our architecture, dhSegment being slightly better 
for illustrations and worse for texts) while using much less parameters (14.4M versus 32.9M 
for dhSegment). This is important as we found that high-resolution images and upscaling data 
augmentation when fine-tuning were crucial to obtain results on par with state-of-the-art 
baseline detection methods.

\section{Conclusion}

To the best of our knowledge, we presented the first robust off-the-shelf system for generic 
element extraction in historical documents. Our approach relies on a single network and 
simple post-processing that simultaneously perform text line and illustration segmentation.  
Its success is based on two key components we introduced: (i) a rich, fast and modular 
synthetic document generation engine and (ii) an adapted segmentation network that predicts 
bounding shapes for illustrations and x-height+border representation for text lines. We 
demonstrated our off-the-shelf approach provides, without any fine-tuning, remarkable 
performances across a wide variety of challenging datasets. Furthermore, when annotated 
training images are available, our network can be used as a good initialization for 
fine-tuning and leads to results on par with the more complex state-of-the-art approaches.

We see our work as a first step toward the development of universal off-the-shelf 
open-sourced methods for practical historical document analysis. Indeed, a lot of efforts has 
been dedicated to boosting performances on specialized challenging datasets. Yet, we believe 
that generic approaches that do not rely on specific trainings for each type of document and 
task are also an important challenge and can have a strong impact to increase applications in 
the humanities. We also think our synthetic generation engine will be easy to improve on by 
adding new elements and more advanced augmentations for even greater generalization 
capabilities.

\section*{Acknowledgments}

This work was supported in part by ANR project EnHerit ANR-17-CE23-0008, project Rapid
Tabasco and gifts from Adobe. We thank Béatrice Joyeux-Prunel, K. Bender, Joanna Fronska, 
Matthieu Husson, Stavros Lazaris, Galla Topalian, Claudia Rabel, Jean-Philippe Moreux and 
Alexandre Turc for their help in the data collection and fruitful discussions. We also thank 
François Darmon, Pierre-Guillaume Raverdy, Tristan Dot and Ryad Kaoua for code testing and 
feedbacks.

\bibliographystyle{IEEEtran}
\bibliography{IEEEabrv,references}

\begin{thebibliography}{10}
\providecommand{\url}[1]{#1}
\csname url@samestyle\endcsname
\providecommand{\newblock}{\relax}
\providecommand{\bibinfo}[2]{#2}
\providecommand{\BIBentrySTDinterwordspacing}{\spaceskip=0pt\relax}
\providecommand{\BIBentryALTinterwordstretchfactor}{4}
\providecommand{\BIBentryALTinterwordspacing}{\spaceskip=\fontdimen2\font plus
\BIBentryALTinterwordstretchfactor\fontdimen3\font minus
  \fontdimen4\font\relax}
\providecommand{\BIBforeignlanguage}[2]{{%
\expandafter\ifx\csname l@#1\endcsname\relax
\typeout{** WARNING: IEEEtran.bst: No hyphenation pattern has been}%
\typeout{** loaded for the language `#1'. Using the pattern for}%
\typeout{** the default language instead.}%
\else
\language=\csname l@#1\endcsname
\fi
#2}}
\providecommand{\BIBdecl}{\relax}
\BIBdecl

\bibitem{gaoICDAR2017CompetitionPage2017}
L.~Gao, X.~Yi, Z.~Jiang, L.~Hao, and Z.~Tang, ``{{ICDAR2017 Competition}} on
  {{Page Object Detection}},'' in \emph{ICDAR}, 2017.

\bibitem{simistiraICDAR2017CompetitionLayout2017}
F.~Simistira, M.~Bouillon, M.~Seuret, M.~Wursch, M.~Alberti, R.~Ingold, and
  M.~Liwicki, ``\BIBforeignlanguage{en}{{{ICDAR2017 Competition}} on {{Layout
  Analysis}} for {{Challenging Medieval Manuscripts}}},'' in
  \emph{\BIBforeignlanguage{en}{ICDAR}}, 2017.

\bibitem{clausnerICDAR2019CompetitionRecognition2019a}
C.~Clausner, A.~Antonacopoulos, and S.~Pletschacher, ``{{ICDAR2019
  Competition}} on {{Recognition}} of {{Documents}} with {{Complex Layouts}} -
  {{RDCL2019}},'' in \emph{ICDAR}, 2019.

\bibitem{clausnerICDAR2019CompetitionRecognition2019}
C.~Clausner, A.~Antonacopoulos, T.~Derrick, and S.~Pletschacher, ``{{ICDAR2019
  Competition}} on {{Recognition}} of {{Early Indian Printed Documents}}
  \textendash{} {{REID2019}},'' in \emph{ICDAR}, 2019.

\bibitem{antonacopoulosRealisticDatasetPerformance2009}
A.~Antonacopoulos, D.~Bridson, C.~Papadopoulos, and S.~Pletschacher,
  ``\BIBforeignlanguage{en}{A {{Realistic Dataset}} for {{Performance
  Evaluation}} of {{Document Layout Analysis}}},'' in
  \emph{\BIBforeignlanguage{en}{ICDAR}}, 2009.

\bibitem{clausnerENPImageGround2015}
C.~Clausner, C.~Papadopoulos, S.~Pletschacher, and A.~Antonacopoulos, ``The
  {{ENP}} image and ground truth dataset of historical newspapers,'' in
  \emph{ICDAR}, 2015.

\bibitem{yangLearningExtractSemantic2017}
X.~Yang, E.~Yumer, P.~Asente, M.~Kraley, D.~Kifer, and C.~L. Giles,
  ``\BIBforeignlanguage{en}{Learning to {{Extract Semantic Structure}} from
  {{Documents Using Multimodal Fully Convolutional Neural Networks}}},'' in
  \emph{\BIBforeignlanguage{en}{CVPR}}, 2017.

\bibitem{zhongPubLayNetLargestDataset2019}
X.~Zhong, J.~Tang, and A.~Jimeno~Yepes, ``{{PubLayNet}}: {{Largest Dataset
  Ever}} for {{Document Layout Analysis}},'' in \emph{ICDAR}, 2019.

\bibitem{diemCBADICDAR2017Competition2017}
M.~Diem, F.~Kleber, S.~Fiel, T.~Gruning, and B.~Gatos,
  ``\BIBforeignlanguage{en}{{{cBAD}}: {{ICDAR2017 Competition}} on {{Baseline
  Detection}}},'' in \emph{\BIBforeignlanguage{en}{ICDAR}}, 2017.

\bibitem{clausnerICFHR2018Competition2018}
C.~Clausner, A.~Antonacopoulos, N.~Mcgregor, and D.~{Wilson-Nunn}, ``{{ICFHR}}
  2018 {{Competition}} on {{Recognition}} of {{Historical Arabic Scientific
  Manuscripts}} \textendash{} {{RASM2018}},'' in \emph{ICFHR}, 2018.

\bibitem{diemCBADICDAR2019Competition2019}
M.~Diem, F.~Kleber, R.~Sablatnig, and B.~Gatos, ``{{cBAD}}: {{ICDAR2019
  Competition}} on {{Baseline Detection}},'' in \emph{ICDAR}, 2019.

\bibitem{rentonHandwrittenTextLine2017}
G.~Renton, C.~Chatelain, S.~Adam, C.~Kermorvant, and T.~Paquet,
  ``\BIBforeignlanguage{en}{Handwritten {{Text Line Segmentation Using Fully
  Convolutional Network}}},'' in \emph{\BIBforeignlanguage{en}{ICDAR}}, 2017.

\bibitem{finkBaselineDetectionHistorical2018}
M.~Fink, T.~Layer, G.~Mackenbrock, and M.~Sprinzl,
  ``\BIBforeignlanguage{en}{Baseline {{Detection}} in {{Historical Documents
  Using Convolutional U}}-{{Nets}}},'' in \emph{\BIBforeignlanguage{en}{DAS}},
  2018.

\bibitem{ronnebergerUnetConvolutionalNetworks2015}
O.~Ronneberger, P.~Fischer, and T.~Brox, ``U-net: Convolutional networks for
  biomedical image segmentation,'' in \emph{MICCAI}, 2015.

\bibitem{heDeepResidualLearning2016}
K.~He, X.~Zhang, S.~Ren, and J.~Sun, ``\BIBforeignlanguage{en}{Deep {{Residual
  Learning}} for {{Image Recognition}}},'' in
  \emph{\BIBforeignlanguage{en}{CVPR}}, 2016.

\bibitem{aresoliveiraDhSegmentGenericDeepLearning2018}
S.~Ares~Oliveira, B.~Seguin, and F.~Kaplan,
  ``\BIBforeignlanguage{en}{{{dhSegment}}: {{A Generic Deep}}-{{Learning
  Approach}} for {{Document Segmentation}}},'' in
  \emph{\BIBforeignlanguage{en}{ICFHR}}, 2018.

\bibitem{gruningTwostageMethodText2019}
T.~Gr{\"u}ning, G.~Leifert, T.~Strau{\ss}, J.~Michael, and R.~Labahn,
  ``\BIBforeignlanguage{en}{A two-stage method for text line detection in
  historical documents},'' \emph{\BIBforeignlanguage{en}{IJDAR}}, 2019.

\bibitem{capobiancoDocEmulToolkitGenerate2017}
S.~Capobianco and S.~Marinai, ``{{DocEmul}}: {{A Toolkit}} to {{Generate
  Structured Historical Documents}},'' in \emph{ICDAR}, 2017.

\bibitem{journetDocCreatorNewSoftware2017}
N.~Journet, M.~Visani, B.~Mansencal, K.~{Van-Cuong}, and A.~Billy,
  ``\BIBforeignlanguage{en}{{{DocCreator}}: {{A New Software}} for {{Creating
  Synthetic Ground}}-{{Truthed Document Images}}},''
  \emph{\BIBforeignlanguage{en}{Journal Imaging}}, vol.~3, p.~62, 2017.

\bibitem{guptaLearningReadSpelling2018}
A.~Gupta, A.~Vedaldi, and A.~Zisserman, ``Learning to read by spelling:
  {{Towards}} unsupervised text recognition,'' in \emph{ICVGIP}, 2018.

\bibitem{yuDilatedResidualNetworks2017}
F.~Yu, V.~Koltun, and T.~Funkhouser, ``\BIBforeignlanguage{en}{Dilated
  {{Residual Networks}}},'' in \emph{\BIBforeignlanguage{en}{CVPR}}, 2017.

\bibitem{nohLearningDeconvolutionNetwork2015}
H.~Noh, S.~Hong, and B.~Han, ``\BIBforeignlanguage{en}{Learning {{Deconvolution
  Network}} for {{Semantic Segmentation}}},'' in
  \emph{\BIBforeignlanguage{en}{ICCV}}, 2015.

\bibitem{ulyanovInstanceNormalizationMissing2016}
D.~Ulyanov, A.~Vedaldi, and V.~Lempitsky, ``\BIBforeignlanguage{en}{Instance
  {{Normalization}}: {{The Missing Ingredient}} for {{Fast Stylization}}},''
  \emph{\BIBforeignlanguage{en}{arXiv:1607.08022}}, 2016.

\bibitem{dengImageNetLargescaleHierarchical2009}
J.~Deng, W.~Dong, R.~Socher, L.-J. Li, {Kai Li}, and {Li Fei-Fei},
  ``\BIBforeignlanguage{en}{{{ImageNet}}: {{A}} large-scale hierarchical image
  database},'' in \emph{\BIBforeignlanguage{en}{CVPR}}, 2009.

\bibitem{glorotUnderstandingDifficultyTraining2010}
X.~Glorot and Y.~Bengio, ``\BIBforeignlanguage{en}{Understanding the difficulty
  of training deep feedforward neural networks},'' in
  \emph{\BIBforeignlanguage{en}{AISTATS}}, 2010.

\bibitem{kingmaAdamMethodStochastic2015}
D.~P. Kingma and J.~Ba, ``\BIBforeignlanguage{en}{Adam: {{A Method}} for
  {{Stochastic Optimization}}},'' in \emph{\BIBforeignlanguage{en}{ICLR}},
  2015.

\bibitem{duttaAnnotationSoftwareImages2019}
A.~Dutta and A.~Zisserman, ``\BIBforeignlanguage{en}{The {{VIA Annotation
  Software}} for {{Images}}, {{Audio}} and {{Video}}},''
  \emph{\BIBforeignlanguage{en}{ACMMM}}, 2019.

\bibitem{heMaskRCNN2017}
K.~He, G.~Gkioxari, P.~Doll{\'a}r, and R.~Girshick, ``Mask {{R}}-{{CNN}},'' in
  \emph{ICCV}, 2017.

\end{thebibliography}

\end{document}